\documentclass[sigconf]{acmart}

\usepackage{algorithm}
\usepackage{algorithmic}
\usepackage{amsmath}
\usepackage{graphicx}
\usepackage{multirow}
\usepackage{booktabs}
\usepackage{subcaption}
\usepackage{float}
\usepackage{bm}
\usepackage{listings}
\usepackage{xcolor}
\usepackage{caption}
\usepackage{enumitem}

\definecolor{codegray}{rgb}{0.99, 0.99, 0.99}
\definecolor{codegreen}{HTML}{089900}

\lstdefinestyle{mystyle}{
    backgroundcolor=\color{codegray},   
    commentstyle=\color{codegreen},
    keywordstyle=\color{blue},
    numberstyle=\tiny\color{gray},
    stringstyle=\color{purple},
    basicstyle=\ttfamily\footnotesize,
    breakatwhitespace=false,         
    breaklines=true,                 
    captionpos=b,                    
    keepspaces=true,                 
    numbers=none,                    
    numbersep=8pt,                  
    showspaces=false,                
    showstringspaces=false,
    showtabs=false,                  
    tabsize=2,
    frame=single,                   
    framesep=1pt,
    rulecolor=\color{black},
    framerule=1pt,
}

\AtBeginDocument{%
  \providecommand\BibTeX{{%
    \normalfont B\kern-0.5em{\scshape i\kern-0.25em b}\kern-0.8em\TeX}}}

\copyrightyear{2024}
\acmYear{2024}
\setcopyright{acmlicensed}\acmConference[GECCO '24]{Genetic and Evolutionary Computation Conference}{July 14--18, 2024}{Melbourne, VIC, Australia}
\acmBooktitle{Genetic and Evolutionary Computation Conference (GECCO '24), July 14--18, 2024, Melbourne, VIC, Australia}
\acmDOI{10.1145/3638529.3654210}
\acmISBN{979-8-4007-0494-9/24/07}

\begin{document}

\title{Tensorized NeuroEvolution of Augmenting Topologies \\ for GPU Acceleration}

\author{Lishuang Wang}
\orcid{0009-0009-2374-8699}
\email{wanglishuang22@gmail.com}
\affiliation{%
  \institution{Southern University of Science and Technology}
  \city{Shenzhen}
  \state{Guangdong}
  \country{China}
  \postcode{518055}
}

\author{Mengfei Zhao}
\orcid{0009-0002-1653-8553}
\email{12211819@mail.sustech.edu.cn}
\affiliation{%
  \institution{Southern University of Science and Technology}
  \city{Shenzhen}
  \state{Guangdong}
  \country{China}
  \postcode{518055}
}

\author{Enyu Liu}
\orcid{0009-0008-5645-8578}
\email{12210417@mail.sustech.edu.cn}
\affiliation{%
  \institution{Southern University of Science and Technology}
  \city{Shenzhen}
  \state{Guangdong}
  \country{China}
  \postcode{518055}
}

\author{Kebin Sun}
\orcid{0009-0008-9213-7835}
\email{sunkebin.cn@gmail.com}
\affiliation{%
  \institution{Southern University of Science and Technology}
  \city{Shenzhen}
  \state{Guangdong}
  \country{China}
  \postcode{518055}
}

\author{Ran Cheng}
\orcid{0000-0001-9410-8263}
\authornote{Corresponding Author}
\email{ranchengcn@gmail.com}
\affiliation{%
    \institution{Southern University of Science and Technology}
    \streetaddress{1088 Xueyuan Avenue}
    \city{Shenzhen}
    \state{Guangdong}
    \country{China}
    \postcode{518055}
}

\begin{abstract}
  The NeuroEvolution of Augmenting Topologies (NEAT) algorithm has received considerable recognition in the field of neuroevolution. 
  Its effectiveness is derived from initiating with simple networks and incrementally evolving both their topologies and weights.
  Although its capability across various challenges is evident, the algorithm's computational efficiency remains an impediment, limiting its scalability potential. 
  In response, this paper introduces a tensorization method for the NEAT algorithm, enabling the transformation of its diverse network topologies and associated operations into uniformly shaped tensors for computation.
  This advancement facilitates the execution of the NEAT algorithm in a parallelized manner across the entire population. 
  Furthermore, we develop TensorNEAT, a library that implements the tensorized NEAT algorithm and its variants, such as CPPN and HyperNEAT. 
  Building upon JAX, TensorNEAT promotes efficient parallel computations via automated function vectorization and hardware acceleration.
  Moreover, the TensorNEAT library supports various benchmark environments including Gym, Brax, and gymnax.
  Through evaluations across a spectrum of robotics control environments in Brax, TensorNEAT achieves up to 500x speedups compared to the existing implementations such as NEAT-Python. 
  Source codes are available at: \url{https://github.com/EMI-Group/tensorneat}.
\end{abstract}




\maketitle

\section{Introduction}
Neuroevolution has emerged as a distinct branch within the field of artificial intelligence (AI). 
Unlike the common approach in machine learning that uses stochastic gradient descent, neuroevolution employs evolutionary algorithms for network optimization. 
Beyond the traditional limit of parameter optimization, it also involves improving elements such as activation functions, hyperparameters, and the overall network structure. 
Moreover, compared to standard machine learning methods which often converge to a single solution, neuroevolution continually promotes a varied set of solutions throughout its exploration \cite{stanley2019designing}.
These defining attributes not only give neuroevolution a strong sense of novelty and diversity but also empower it for certain open-ended challenges \cite{lehman2011evolving, mouret2015illuminating}.

The NeuroEvolution of Augmenting Topologies (NEAT) \cite{stanley2002evolving} is well-recognized in the neuroevolution literature. 
Since its introduction in 2002, NEAT has been shown to be useful in various areas such as game AI \cite{stanley2006real, pham2018playing}, robotics \cite{silva2012odneat, auerbach2011evolving} and self-driving systems \cite{yuksel2018agent}.
The original NEAT algorithm provides a basic framework, and the subsequent works have kept exploiting its potential.
For instance, offshoots like HyperNEAT \cite{stanley2009hypercube} and ES-HyperNEAT \cite{risi2010evolving} adopted indirect encoding for expansive networks;
DeepNEAT and CoDeepNEAT \cite{miikkulainen2019evolving} combined gradient descent methods to delve deeper into neural architecture. 
Recently, RankNEAT \cite{rankneat} used neuroevolution for preference learning tasks, effectively optimizing network architectures in subjectively labeled data environments.
Such ongoing development and flexibility highlight NEAT's continued importance and appeal in the field.

During the past years, GPU acceleration has been a driving force behind the rapid advancements in AI, especially in the field of deep learning. 
The advancement in such hardware acceleration has facilitated the expansion of deep learning, with modern large language models incorporating hundreds of billions of parameters \cite{brown2020language}. 
Using GPUs for tasks like inference and back-propagation has led to significant speed improvements.
Meanwhile, within the field of neuroevolution, there is also a growing interest in leveraging GPUs for hardware acceleration. 
Building upon the JAX framework~\cite{frostig2018compiling}, some pioneering works such as EvoJAX \cite{tang2022evojax}, evosax \cite{lange2022evosax}, and EvoX \cite{huang2023evox} represent this trend, seeking to tap into the robust computational capabilities of GPU to reduce the runtime of neuroevolution algorithms, especially when handling large population sizes or extensive problem scales. 

However, despite the rapid emergence of these GPU-accelerated libraries, NEAT is left behind. 
Our analysis showed that though the NEAT algorithm has been implemented in various programming languages~\cite{McIntyre_neat-python, MultiNEAT, MonopolyNEAT}, few of them harness GPUs to boost their performance. 
On those rare occasions when GPUs are used~\cite{pytorchneat}, the focus has primarily been on accelerating the network inference process of the NEAT algorithm, while other crucial parts, such as network search processes, are often overlooked. Moreover, they do not efficiently utilize GPU parallel computing, as key operations like fitness evaluation and mutation are executed sequentially in their frameworks.
This limitation can be attributed to the unique nature of NEAT: It employs networks with \emph{continuously evolving topologies} during the algorithm's execution. This characteristic poses a challenge for efficient GPU implementation.

To bridge this gap, we develop TensorNEAT, a tensorized NEAT library optimized for GPU acceleration.
TensorNEAT employs a new tensorization method that transforms networks of varying topologies into tensors with uniform shape, ensuring that operations in the NEAT algorithm can be executed in parallel across the entire population.
Implemented within the JAX framework, TensorNEAT enables automatic GPU acceleration without any specific configuration. 
In comparison with the existing popular open-source implementation of the NEAT algorithm, TensorNEAT achieves up to 500x speedups. Overall, our contributions are summarized as follows.

\begin{itemize}[topsep=0pt,partopsep=0pt]
    \item We propose a tensorization method, which enables the transformation of networks with various topologies and their associated operations in the NEAT algorithm into uniformly structured tensors for tensor computation. This method allows operations within the NEAT algorithm to be executed in parallel across the entire population, thereby enhancing the efficiency of the process.

    \item We develop TensorNEAT, a GPU-accelerated NEAT library based on JAX, characterized by high efficiency, flexible adaptability, and rich capabilities. TensorNEAT supports full GPU acceleration of representative NEAT algorithms, including the original NEAT algorithm, CPPN~\cite{stanley2007compositional}, and HyperNEAT~\cite{stanley2009hypercube}. It also provides seamless interfaces with advanced control benchmarks, including Brax~\cite{freeman2021brax} and Gymnax~\cite{lange2022gymnax}, featuring GPU-accelerated environments with various classical control and robotics control tasks.

    \item We assessed TensorNEAT's performance in a spectrum of complex robotics control tasks, benchmarked against the NEAT-Python library~\cite{McIntyre_neat-python}. The results show that TensorNEAT significantly outperforms in terms of execution speed, especially under high computational demands and across various population sizes and network scales.
\end{itemize}

\section{Background}

\subsection{NeuroEvolution of Augmenting Topologies}

Introduced by Kenneth O. Stanley and Risto Miikkulainen in 2002, the NeuroEvolution of Augmenting Topologies (NEAT) algorithm \cite{stanley2002evolving} represents a novel approach in neuroevolution. 
The NEAT algorithm manages a range of neural networks and simultaneously optimizes their topologies and weights to identify the most effective networks tailored for designated tasks. Algorithm~\ref{alg:neat} in appendix~\ref{Appendix_neat} outlines NEAT's core procedure. Starting with a population of simple neural networks, it iterates through evolutionary cycles of species formation, fitness evaluation, and genetic operations. This process dynamically refines the networks until it achieves desired fitness levels or reaches a generational cap, ultimately yielding the optimal network structure.

Setting itself apart from alternative neuroevolution algorithms, NEAT employs three distinctive techniques:

\begin{itemize}[topsep=0pt,partopsep=0pt]
    \item \textbf{Incremental Topological Expansion}: The NEAT algorithm initializes its evolutionary search with the most basic networks, consisting of just a single hidden node bridging inputs to outputs. As evolution advances, the algorithm incorporates new nodes and connections, progressively sophisticating the network topology. This strategy pragmatically narrows down the search space, facilitating the resolution of complex challenges using more streamlined network structures.
    \item \textbf{Historical Markers for Nodes}: Each node in a NEAT network is tagged with a unique historical marker. During the crossover process in the NEAT algorithm, only nodes with identical markers are combined. This method adeptly navigates the challenges of combining networks that possess different topological configurations.
    \item \textbf{Species-based Population Segmentation}: NEAT categorizes its entire population into individual species. Conventional genetic procedures, such as selection, mutation, and crossover, are executed independently within each species. This approach serves dual purposes: it not only shields potentially advantageous network structures from premature extinction, but also amplifies the diversity of solutions within the evolutionary exploration.
\end{itemize}

\begin{figure}[b]
    \centering
    \includegraphics[width=0.9\columnwidth]{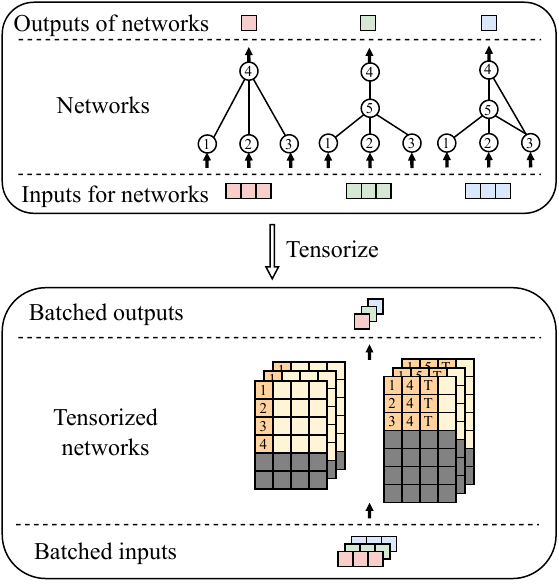}
    \caption{Illustration of the fundamental acceleration principle underlying our tensorization method. In traditional network computations, each network individually processes its inputs. 
    By contrast, with our tensorization method, a single computation suffices to derive the batched output for all networks.
    }
    \label{fig:pop_forward}
\end{figure}

Since its inception, the NEAT algorithm has undergone significant evolution, leading to the emergence of several innovative variants, each addressing unique challenges and applications. HyperNEAT~\cite{stanley2009hypercube} marked a pivotal advancement by employing an indirect encoding scheme. This approach allows efficient handling of large-scale neural networks, leveraging geometric regularities to create complex network topologies that can be applied to tasks requiring substantial representational capacity. Building upon this concept, ES-HyperNEAT~\cite{risi2010evolving} further refined the approach, introducing enhanced techniques for evolving network structures with even greater scalability and adaptability.

Then the development of DeepNEAT and CoDeepNEAT~\cite{miikkulainen2019evolving} represented a significant leap in integrating neuroevolution with deep learning principles. DeepNEAT extended the NEAT algorithm to the realm of deep neural networks. It leverages gradient descent to explore and optimize complex, layered neural architectures. CoDeepNEAT expanded this concept further, introducing a co-evolutionary approach that allowed for the simultaneous evolution of both the topology and components of deep neural networks. This fusion of gradient descent and evolutionary strategies enabled a more thorough and nuanced exploration of neural structures, opening new avenues in areas like feature learning and hierarchical network construction. 

RankNEAT~\cite{rankneat} is another variant of the NEAT algorithm. It overcomes the limitations of Stochastic Gradient Descent by using neuroevolution to optimize network architectures, effectively reducing overfitting. RankNEAT has proven to be effective in affective computing, outshining traditional methods like RankNet. It is particularly successful in analyzing player arousal from game footage, highlighting its potential in handling subjective data.

Together, these variants of NEAT demonstrate the superior algorithm's versatility and its continuous adaptation to address the evolving complexities of neural network design and application.

\begin{figure*}[tbp]
    \centering
    \includegraphics[width=0.9\textwidth]{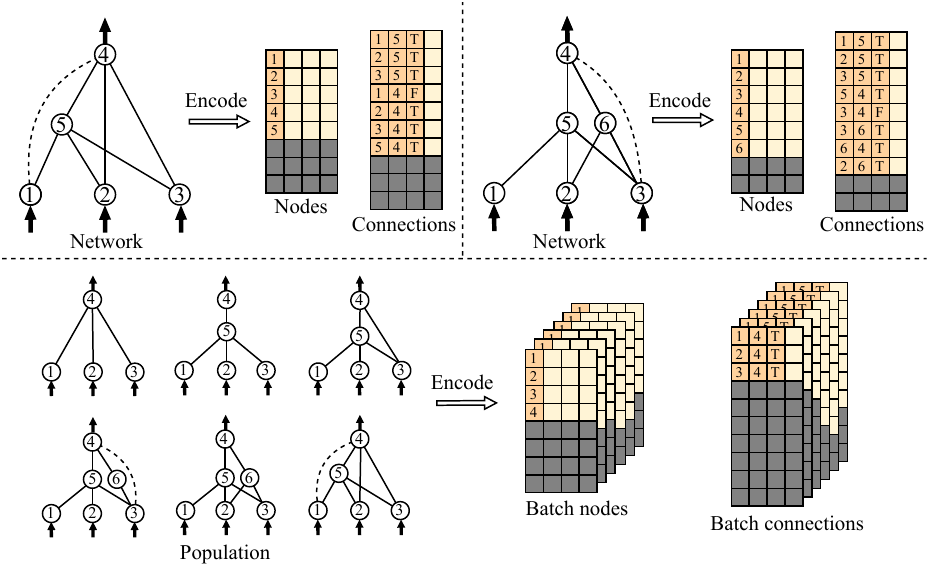}
    \caption{Illustration of the network encoding process. 
    In our method, networks with varying topological structures are transformed into uniformly shaped tensors, enabling the representation of the entire network population as batched tensors. 
    The orange cells symbolize attributes specific to the NEAT algorithm, such as historical markers and enabled flags. 
    The yellow cells denote the attributes of the network's nodes and connections, including biases, weights, and activation functions. 
    The gray cells represent sections filled with \texttt{NaN} to ensure consistent tensor shapes.}
    \label{fig:pop_encoding}
\end{figure*} 

\subsection{NEAT Libraries}
In the past two decades, the research community has witnessed the emergence of various NEAT libraries, including NEAT-Python~\cite{McIntyre_neat-python}, MultiNEAT~\cite{MultiNEAT}, and MonopolyNEAT~\cite{MonopolyNEAT}. 
Among these libraries, the NEAT-Python library \cite{McIntyre_neat-python} stands out as the foremost open-source implementation of NEAT at present. With over 1,200 GitHub stars, it has served as the foundational base for numerous academic research projects related to NEAT \cite{rankneat, gao2021neat, sarti2021neat}.
These NEAT implementations use the object-oriented programming paradigm, involving the representation of core NEAT components such as populations, species, genomes, and genes as objects. 
This programming paradigm provides remarkable transparency, enhancing code readability and thereby enabling those new to the domain to rapidly grasp the complexities of the NEAT algorithm. 
However, maintaining objects also brings extra computational overheads in the running process, especially when encountered with large-scale problems or substantial population sizes. 

Most existing NEAT implementations do not utilize GPUs to accelerate computation, with a few exceptions such as PyTorchNEAT~\cite{pytorchneat}. 
These libraries integrate tensor deep learning, such as PyTorch~\cite{pytorch} and TensorFlow~\cite{tensorflow}, enabling networks generated by NEAT to perform inference on GPUs. 
This approach also enables batch inference for multiple inputs, thereby enhancing network inference speed when dealing with a large number of input data.
However, these libraries still rely on the object-oriented programming paradigm and have optimized only the inference aspect of the network.
The acceleration of the neuroevolution process, which NEAT uses for network search, has not yet been achieved.
Moreover, although individual network inferences can be batch-processed on GPUs, the inference process for the entire population in the NEAT algorithm remains sequential.
Consequently, existing NEAT implementations cannot fully make use of the high parallel processing capabilities of GPUs.

\section{Tensorization of NEAT}

To overcome the limitations of current NEAT implementations and fully leverage modern hardware to accelerate the NEAT algorithm's efficiency, we introduce a new tensorization method. 
Figure~\ref{fig:pop_forward} depicts the key principle of acceleration supporting this approach. 
Our method facilitates the transformation of various network topologies and their related operations into uniformly structured tensors, suitable for tensor computation. 
By implementing function vectorization within network operations, our approach enables parallel processing across the entire population, which effectively harnesses the parallel computing power of GPUs. 
In the following sections, we detail the specific tensorization techniques employed in network encoding and operations.

\begin{figure}[t]
    \centering
    \includegraphics[width=0.9\columnwidth]{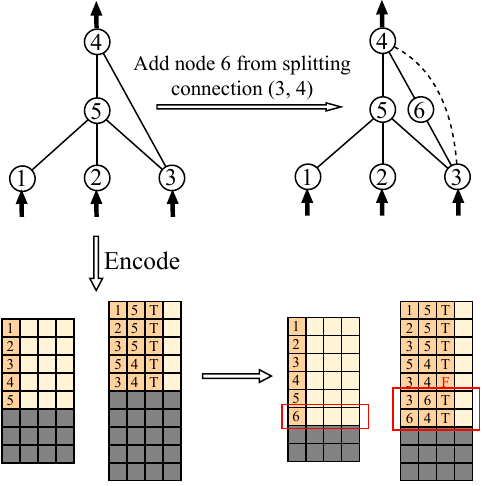}
    \caption{An illustration of tensorized network operations. It demonstrates how traditional network operations are converted into equivalent tensor operations during tensorization. Changes from the original format are highlighted in red, underscoring the modifications made within the tensor.}
    \label{fig:operations}
\end{figure}

\begin{figure*}
    \centering
    \begin{subfigure}{0.43\textwidth}
        \centering
        \includegraphics[width=\textwidth]{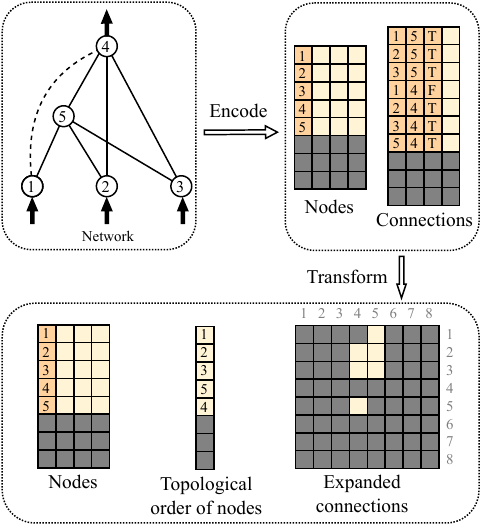}
    \end{subfigure}
    \hfill\vline\hfill
    \begin{subfigure}{0.54\textwidth}
        \centering
        \includegraphics[width=\textwidth]{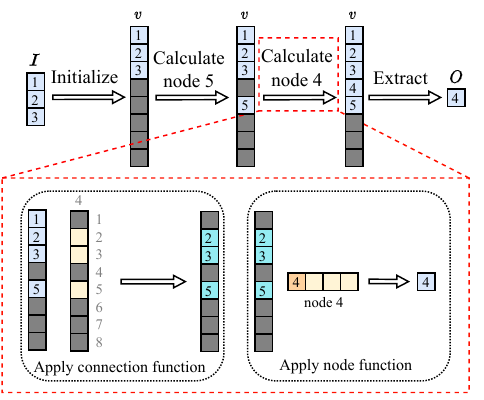}
    \end{subfigure}
    \caption{Illustration of tensorized network inference process. 
    It transforms a feedforward network's topology into tensors and the subsequent calculation of node values for network inference. 
    The network is first encoded into node and connection tensors, which are then ordered and expanded for processing. 
    Finally, values are calculated through connection and node functions to produce the output.}
    \label{fig:transform_and_forward}
\end{figure*}

\subsection{Tensorized Encodings}

In NEAT, a network $\mathcal{N}$ can be represented as:
$$
\mathcal{N} = \langle N, C \rangle,
$$
where $N$, $C$ are nodes and connections in $\mathcal{N}$, respectively. They can be represented as:
$$
N = \{n_1, n_2, n_3, \ldots\} \quad \text{and} \quad C = \{c_1, c_2, c_3, \ldots\},
$$
with $n_i$ denoting the $i$-th node in $N$ and $c_i$ denoting the $i$-th connection in $C$. 

In the NEAT algorithm, each node $n$ can be represented as a tuple consisting of a historical marker and its attributes:
$$ 
n = (k, \text{attr}_1, \text{attr}_2, \ldots),
$$
where $k \in \mathbb{N} $ is the historical marking, and \(\text{attr}_i\) is the $i$-th attribute of \(n\). 
Similarly, a connection $c$ can be represented as:
\[
c = (k_i, k_o, e, \text{attr}_1, \text{attr}_2, \ldots),
\]
where $k_i \in \mathbb{N} $ and $k_o \in \mathbb{N}$ are the historical markings of the input and output nodes of $c$, respectively, $e \in \{\texttt{True}, \texttt{False}\}$ is the enabled flag. The attributes of a node $n$ can be bias $b$, aggregation function $f_{\text{agg}}$ like sum, and activation function $f_{\text{act}}$ like sigmoid and the attributes of a connection $c$ can be weight $w$. Attributes that can be represented numerically, such as \(b\) and \(w\), are directly stored in tensors. Those attributes that are not inherently numerical, such as \(f_{\text{agg}}\) and \(f_{\text{act}}\), are stored in tensors as specific integer values, for example, tanh is represented by 1, and sigmoid by 2.

We can use one-dimensional tensors to encode $n$ and $c$:
\begin{align*}
    \boldsymbol{n} & = [k, \text{attr}_1, \text{attr}_2, \ldots] \in \mathbb{R}^{1 + \texttt{noa}(n)},\\    
    \boldsymbol{c} & = [k_i, k_o, e, \text{attr}_1, \text{attr}_2, \ldots] \in \mathbb{R}^{3 + \texttt{noa}(c)},
\end{align*}
where \(\texttt{noa}(i)\) denotes the number of attributes of \(i\).
 
Then, the sets $N$ and $C$ can be represented as tensors $\boldsymbol{N}$ and $\boldsymbol{C}$, respectively:
\begin{align*}
    \boldsymbol{N} & = [\boldsymbol{n}_1, \boldsymbol{n}_2, \ldots] \in \mathbb{R}^{|N|\times\texttt{len}(\boldsymbol{n})}, \\
    \boldsymbol{C} & = [\boldsymbol{c}_1, \boldsymbol{c}_2, \ldots] \in \mathbb{R}^{|C|\times\texttt{len}(\boldsymbol{c})},
\end{align*}
where $|\cdot|$ is the cardinality of a set, $\texttt{len}(\cdot)$ is the length of a one-dimensional tensor, and $\boldsymbol{n}_i$ and $\boldsymbol{c}_j$ are the tensor representations of the $i$-th node and $j$-th connection, respectively.   

To address the issue where each network in the population possesses a varying number of nodes and connections, we employ tensor padding using \texttt{NaN} values for alignment. We set predefined maximum limits for the number of nodes and connections, represented as $|N|_\text{max}$ and $|C|_\text{max}$, respectively. The resulting padded tensors, $\hat{\boldsymbol{N}}$ and $\hat{\boldsymbol{C}}$, are formulated as:
\begin{align*}
\hat{\boldsymbol{N}}& = [\boldsymbol{n}_1, \boldsymbol{n}_2, \ldots, \texttt{NaN}, \ldots] \in \mathbb{R}^{|N|_\text{max}\times\texttt{len}(\boldsymbol{n})}, \\
\hat{\boldsymbol{C}}& = [\boldsymbol{c}_1, \boldsymbol{c}_2, \ldots, \texttt{NaN}, \ldots] \in \mathbb{R}^{|C|_\text{max}\times\texttt{len}(\boldsymbol{c})}. 
\end{align*} 

Upon alignment, the tensors $\hat{\boldsymbol{N}}$ and $\hat{\boldsymbol{C}}$ of each network in the population can be concatenated, allowing us to express the entire population using two tensors: $\boldsymbol{P}_{N}$ and $\boldsymbol{P}_{C}$. These tensors encompass all nodes and connections within the population, respectively. Formally, the concatenated tensors, $\boldsymbol{P}_{N}$ and $\boldsymbol{P}_{C}$, are defined as:
\begin{align*}
\boldsymbol{P}_{N} & = [\hat{\boldsymbol{N}}_1, \hat{\boldsymbol{N}}_2, \ldots] \in \mathbb{R}^{P \times |N|_\text{max} \times\texttt{len}(\boldsymbol{n})}, \\
\boldsymbol{P}_{C} & = [\hat{\boldsymbol{C}}_1, \hat{\boldsymbol{C}}_2, \ldots] \in \mathbb{R}^{P \times |C|_\text{max} \times\texttt{len}(\boldsymbol{c})},
\end{align*}
where $P$ denotes the population size, $\hat{\boldsymbol{N}}_i$ and $\hat{\boldsymbol{C}}_i$ represent the tensor representations of the node and connection sets of the $i$-th network, respectively.  
Fig.~\ref{fig:pop_encoding} provides a graphical representation of the tensorized encoding process.

\subsection{Tensorized Operations}
Upon encoding networks as tensors, we subsequently express operations on NEAT networks as corresponding tensor operations. In this subsection, we detail the tensorized representations of three fundamental operations: node modification, connection modification, and attribute modification.

\subsubsection{Node Modification}
Given a network \(\mathcal{N} = \langle N, C \rangle\), node set modifications in \(N\) can involve either the addition of a new node \(n\) or the removal of an existing node \(n\):
\begin{align*}
N' = N \cup \{n\} \quad \text{or} \quad N' = N \setminus \{n\}.
\end{align*}

In the tensorized representation $\hat{\boldsymbol{N}} \in \mathbb{R}^{|N|_\text{max}\times\texttt{len}(\boldsymbol{n})} $ of the node set \(N\), the tensorized operation for node addition can be represented as:
\[
\hat{\boldsymbol{N}}[r_i] \leftarrow \boldsymbol{n}_\text{new},
\]
where \(r_i\) denotes the index of the first \texttt{NaN} row in \(\hat{\boldsymbol{N}}\), \(\boldsymbol{n}_\text{new}\) stands for the tensor representation of the node being added, \([\cdot]\) represents tensor slicing, and \(\leftarrow\) indicates the assignment operation.

Conversely, the tensorized operation for node removal can be depicted as:
\[
\hat{\boldsymbol{N}}[r_j] \leftarrow \texttt{NaN},
\]
with \(r_j\) representing the index of the node for removal.

\subsubsection{Connection Modification}
Given a network \(\mathcal{N} = \langle N, C \rangle\), modifications in the connection set \(C\) can entail either adding a new connection or eliminating an existing connection \(c\):
\begin{align*}
    C' = C \cup \{c\} \quad \text{or} \quad C' = C \setminus \{c\}.
\end{align*}

In the tensorized representation, \(\hat{\boldsymbol{C}}\), of \(C\), the operations of connection addition or removal can be depicted as:
\[
    \hat{\boldsymbol{C}}[r_i] \leftarrow \boldsymbol{c}_\text{new} \quad \text{or} \quad 
    \hat{\boldsymbol{C}}[r_j] \leftarrow \texttt{NaN}, 
\]
respectively. Here, \(r_i\) indicates the index of the initial \texttt{NaN} row in \(\hat{\boldsymbol{C}}\), \(\boldsymbol{c}_\text{new}\) represents the tensor form of the connection being introduced, and \(r_j\) signifies the index of the connection for removal.

\subsubsection{Attribute Modification}
NEAT is designed not only to modify the network structures but also the internal attributes, either in a node or a connection. Specifically, with node $n = (k, \text{attr}_1, \text{attr}_2, \ldots)$, 
where $k \in \mathbb{N} $ is the historical marking, and \(\text{attr}_i\) is the $i$-th attribute of $n$, 
when modifying the \(j\)-th attribute in a node, the transformation can be represented as:
\[
n' = (k, \text{attr}_1, \text{attr}_2, \ldots, \text{attr}'_j, \ldots),
\]
and the corresponding tensorized operation is:
\[
\boldsymbol{n}[1 + j] \leftarrow \text{attr}'_j.
\]
Similarly, for a connection $c = (k_i, k_o, e, \text{attr}_1, \text{attr}_2, \ldots)$, where $k_i \in \mathbb{N} $ and $k_o \in \mathbb{N}$ are the historical markings of the input and output nodes of $c$, respectively, $e \in \{\text{True}, \text{False}\}$ is the enabled flag, when modifying the \(j\)-th attribute in a connection, the transformation can represented as:
\[
c' = (k_i, k_o, e, \text{attr}_1, \text{attr}_2, \ldots, \text{attr}'_j, \ldots),
\]
and the corresponding tensorized operation is:
\[
\boldsymbol{c}[3 + j] \leftarrow \text{attr}'_j.
\]

By combining the aforementioned three operations, operations for searching networks in the NEAT algorithm including Mutation and Crossover can be transformed into operations on tensors. Fig.~\ref{fig:operations} provides a graphical representation of the tensorized network operations.

\subsection{Tensorized Network Inference}
Another crucial component in NEAT is the inference process, where a network receives inputs and generates corresponding outputs based on its topologies and weights. For a network $\mathcal{N}$, given its node tensor $\hat{\boldsymbol{N}}$ and connection tensor $\hat{\boldsymbol{C}}$, the inference process can be represented as:
$$
\bm{O} = \text{inference}_{\hat{\boldsymbol{N}}, \hat{\boldsymbol{C}}}(\bm{I}), 
$$
where $\bm{O}$ and $\bm{I}$ denote the outputs and inputs in the inference process, respectively.

In our tensorization method, the inference process is divided into two stages: transformation and calculation. `Transformation' involves converting the node tensor $\hat{\boldsymbol{N}}$ and connection tensor $\hat{\boldsymbol{C}}$ into formats more conducive to network inference. 
`Calculation' refers to computing the output using the tensors produced in the transformation stage.
When a network undergoes multiple inference operations, it only needs to be transformed once. Networks in the NEAT algorithm can be categorized as either feedforward or recurrent, based on the presence or absence of cycles in their topological structure. Here, we primarily focus on the transformation and calculation processes in feedforward networks.

In feedforward networks, the transformation process creates two new tensors: the topological order of nodes $\boldsymbol{N}_{\text{order}} \in \mathbb{R}^{|N|_\text{max}}$ and the expanded connections $\boldsymbol{C}_{\text{exp}} \in \mathbb{R}^{|N|_\text{max} \times |N|_\text{max} \times \texttt{noa}(c)}$, where $|N|_\text{max}$ represents the predefined maximum limit for the number of nodes and $\texttt{noa}(c)$ denoting the number of attributes of connections in the network.

Given the absence of cycles in the network, topological sorting is employed to obtain $\boldsymbol{N}_{\text{order}}$. For $\boldsymbol{C}_{\text{exp}}$, the generation rule can be represented as:
$$
    \boldsymbol{C}_{\text{exp}}[\hat{\boldsymbol{C}}[i][0, 1]] \leftarrow 
    \begin{cases}
        \texttt{NaN}, & \hat{\boldsymbol{C}}[i][2] = 0\\
        \hat{\boldsymbol{C}}[i][2:], & \hat{\boldsymbol{C}}[i][2] = 1
    \end{cases},
$$
where $i = 0, 1, 2, \ldots, |C|_\text{max}$, and $|C|_\text{max}$ is the predefined maximum limit for the number of connections. 
Recall that in each line $c$ of $\hat{\boldsymbol{C}}$, the values are $(k_i, k_o, e, \text{attr}1, \text{attr}2, \ldots)$, with $c[0]=k_i$ and $c[1]=k_o$ indicating the indices of the input and output nodes of the connection, respectively, and $c[2]=e \in \{\texttt{True}, \texttt{False}\}$ representing the enabled flag.
Locations not updated by this rule default to the value $\texttt{NaN}$. The tensors $\hat{\boldsymbol{N}}$, $\boldsymbol{N}_{\text{order}}$, and $\boldsymbol{C}_{\text{exp}}$ are then used as inputs for the calculation process.

In the forward process, we utilize the transformed tensors $\hat{\boldsymbol{N}}$, $\boldsymbol{N}_{\text{order}}$, $\boldsymbol{C}_{\text{exp}}$, and the input $\bm{I}$ to calculate the output $\bm{O}$. We maintain a tensor $\boldsymbol{v} \in \mathbb{R}^{|N|_\text{max}}$ to store the values of nodes in the network. Initially, $\boldsymbol{v}$ is set to the default value $\texttt{NaN}$ and then updated with:
$$
\boldsymbol{v}[k_{\text{input}}] \leftarrow \bm{I},
$$
where $k_{\text{input}}$ denotes the indices of input nodes in the network.
The value of nodes is iteratively calculated in the order specified by $\boldsymbol{N}_{\text{order}}$.
The rule to obtain the value $\boldsymbol{v}[k]$ of node $\boldsymbol{n}_k$ can be expressed as:
\begin{align*}
\boldsymbol{v}[k] \leftarrow f_n(f_c(\boldsymbol{v} \ |\  \boldsymbol{C}_{\text{exp}}[:][k]) \ | \ \hat{\boldsymbol{N}}[k][1:]),
\end{align*}
where $\boldsymbol{C}_{\text{exp}}[:][k]$ indicates the attributes of all connections to $\boldsymbol{n}_k$, and $\hat{\boldsymbol{N}}[k][1:]$ denotes the attributes of $\boldsymbol{n}_k$. $f_c$ and $f_n$ are the calculation functions for connections and nodes in the network, respectively. For a network with connection attribute weight $w$, and node attributes bias $b$, aggregation function $f_{\text{agg}}$ and activation function $f_{\text{act}}$, $f_c$ and $f_n$ can be defined as:
\begin{align*}
    f_c(\bm{I}_{\text{conn}}\ |\ w) &= w\bm{I}_{\text{conn}},\\
    f_n(\bm{I}_{\text{node}}\ |\ b, f_{\text{agg}}, f_{\text{act}}) &= f_{\text{act}}(f_{\text{agg}}(\bm{I}_{\text{node}}) + b),
\end{align*}
where $\bm{I}_{\text{conn}}$ and $\bm{I}_{\text{nodes}}$ denote the inputs for connections and nodes, respectively.

After computing the values of all nodes, the tensor $\boldsymbol{v}[k_{\text{output}}]$ represents the network's output, with $k_{\text{output}}$ indicating the indices of output nodes in the network.
Fig.~\ref{fig:transform_and_forward} graphically depicts the tensorized network inference process.

\section{Implementation of TensorNEAT}


\subsection{JAX-based Hardware Acceleration}
JAX \cite{frostig2018compiling} is an open-source numerical computing library, offering APIs similar to NumPy \cite{harris2020array} and enabling efficient execution across various hardware platforms (CPU/GPU/TPU). 
Leveraging the capabilities of XLA, JAX facilitates the transformation of numerical code into optimized machine instructions. 
The optimization techniques provided by JAX have supported the development of numerous projects in evolutionary computation, as evidenced by \cite{tang2022evojax}, \cite{lange2022evosax}, \cite{lim2022qdax}, and \cite{huang2023evox}. 
These advancements significantly contribute to JAX's growing popularity in the scientific community.

TensorNEAT, integrating JAX, utilizes the functional programming paradigm to implement our proposed tensorization methods.
This integration allows NEAT to effectively use hardware accelerators such as GPUs and TPUs. 
With uniform tensor shapes in network encoding, several NEAT operations such as mutation, crossover, and network inference can be vectorized across the population dimension.
This vectorization, leveraging \texttt{jax.vmap} and \texttt{jax.pmap} functions, is suitable for both single and multi-device configurations.

\subsection{User-friendly Interfaces}
Designed with user-friendly interfaces, TensorNEAT provides mechanisms for adjusting algorithms to specific requirements.
It offers an extensive set of hyperparameters, allowing users to fine-tune various computational elements of the NEAT algorithm. 
Additionally, its modular problem templates facilitate the integration of specialized problems. 
TensorNEAT also offers several open interfaces.
By implementing a few functions, users can define the behavior of networks within the NEAT algorithm.
A notable feature is the interface supporting the evolution of advanced network architectures, including Spiking Neural Networks \cite{ghosh2009spiking} and Binary Neural Networks \cite{hubara2016binarized}. 
Details on the hyperparameters and the interfaces are elaborated in Appendix~\ref{Appendix_a} and Appendix~\ref{Appendix_b}.

\subsection{Feature-rich Extensions}
Extending beyond the conventional NEAT paradigm, TensorNEAT includes notable algorithmic extensions such as Compositional Pattern Producing Networks (CPPN) \cite{stanley2007compositional} and HyperNEAT \cite{stanley2009hypercube}, tailored for parallel processing on hardware accelerators.
For evaluation, TensorNEAT provides a suite of standard test benchmarks, covering areas from numerical optimization to function approximation.
Moreover, it integrates seamlessly with leading reinforcement learning environments like Gym \cite{brockman2016openai}, and hardware-optimized platforms such as gymnax \cite{lange2022gymnax} and Brax \cite{freeman2021brax}, thus allowing users to evaluate the performance of the NEAT algorithm in various settings.

\section{Experiment}

This section presents an empirical comparison between TensorNEAT and NEAT-Python, focusing on their performance in three robotics control tasks: Swimmer, Hopper, and Halfcheetah, within the Brax environment. 
The experiments were conducted with both TensorNEAT and NEAT-Python configured using uniform parameter settings. 
Data presented are the average outcomes from ten independent trials, complete with 95\% confidence intervals to ensure statistical robustness. Details on the experiment settings are shown in Appendix~\ref{Appendix_c}.

\begin{table*}[htbp]
    \centering
        \caption{Runtimes over $100$ generations on different hardware configurations, with a constant population size of $10,000$.}
    \begin{tabular}{c|c|c|c|c}
            
        \hline
        Task & Framework & Hardware & Time ($s$) &Speedup\\
        \hline
        
        \multirow{5}{*}{Swimmer} & NEAT-Python & EPYC 7543 (CPU) & $42279.63 \pm 3031.97$ & $1.00$ \\
        
        \cline{2-5}
        & \multirow{4}{*}{TensorNEAT} & RTX 4090 & $\bm{215.70 \pm 6.35}$ & $\bm{196.01}$\\  
                                    && RTX 3090 & $292.22 \pm 20.19$ & $144.68$\\
                                    && RTX 2080Ti & $434.94 \pm 12.49$ & $97.21$\\
                                    && EPYC 7543 (CPU) & $14678.10 \pm 616.85$ & $2.88$\\

        \hline
        \multirow{5}{*}{Hopper} & NEAT-Python & EPYC 7543 (CPU) & $14438.13 \pm 900.68$ & $1.00$\\
        
        \cline{2-5}
        & \multirow{4}{*}{TensorNEAT} & RTX 4090 & $\bm{241.30 \pm 9.41}$ & $\bm{59.83}$\\  
                                    && RTX 3090 & $336.08 \pm 26.96$ & $42.96$\\
                                    && RTX 2080Ti & $518.40 \pm 7.22$ & $27.85$\\
                                    && EPYC 7543 (CPU) & $13473.01 \pm 544.07$ & $1.07$\\
                                    
        \hline
        
        \multirow{5}{*}{Halfcheetah} & NEAT-Python & EPYC 7543 (CPU) & $149516.00 \pm 4817.90$ & $1.00$\\
        \cline{2-5}
        & \multirow{4}{*}{TensorNEAT} & RTX 4090 & $\bm{274.74 \pm 14.21}$ &  $\bm{544.21}$\\  
                                    && RTX 3090 & $487.82 \pm 19.05$ & $306.50$\\
                                    && RTX 2080Ti & $705.13 \pm 16.02$ &  $212.04$\\
                                    && EPYC 7543 (CPU) & $15914.08 \pm 4005.08$ & $9.40$\\

        \hline
    \end{tabular}
    \label{table:time_hardware}
\end{table*}

\begin{figure}[t]
    \centering
    \begin{subfigure}{0.49\columnwidth}
        \centering
        \includegraphics[width=\columnwidth]{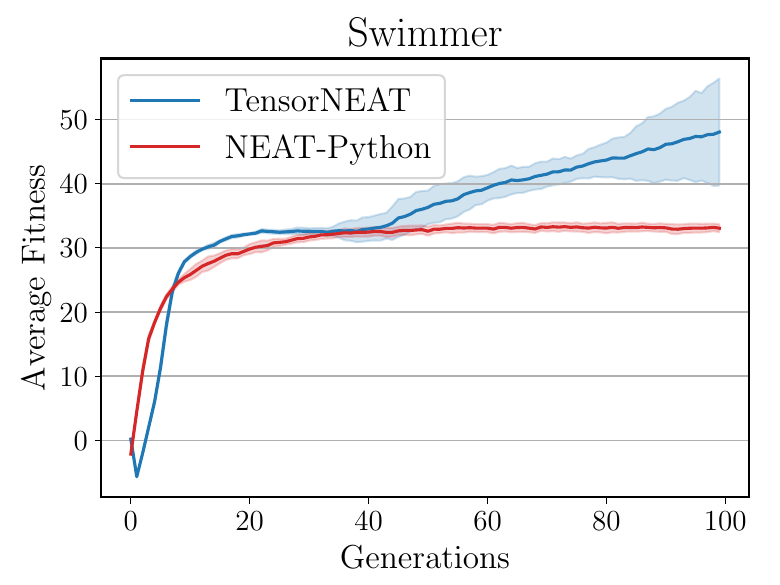}
    \end{subfigure}
    \hfill
    \begin{subfigure}{0.49\columnwidth}
        \centering
        \includegraphics[width=\columnwidth]{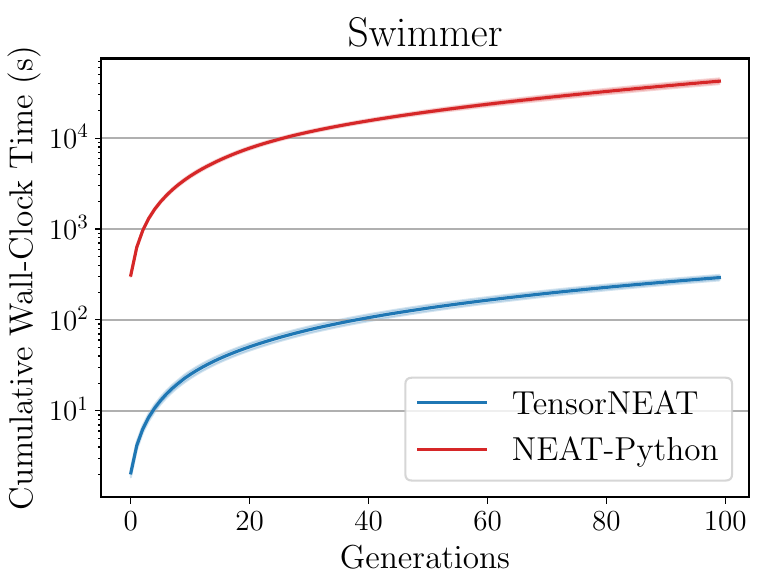}
    \end{subfigure}
    \hfill
    \begin{subfigure}{0.49\columnwidth}
        \centering
        \includegraphics[width=\columnwidth]{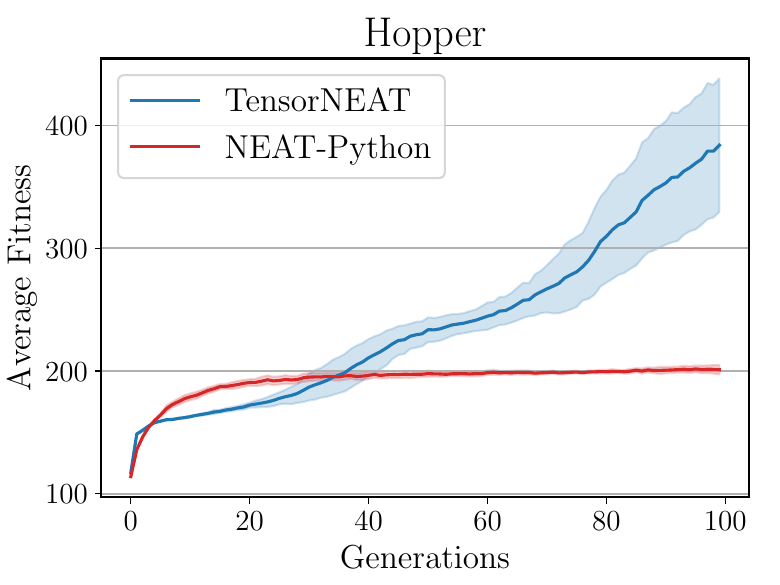}
    \end{subfigure}
    \hfill
    \begin{subfigure}{0.49\columnwidth}
        \centering
        \includegraphics[width=\columnwidth]{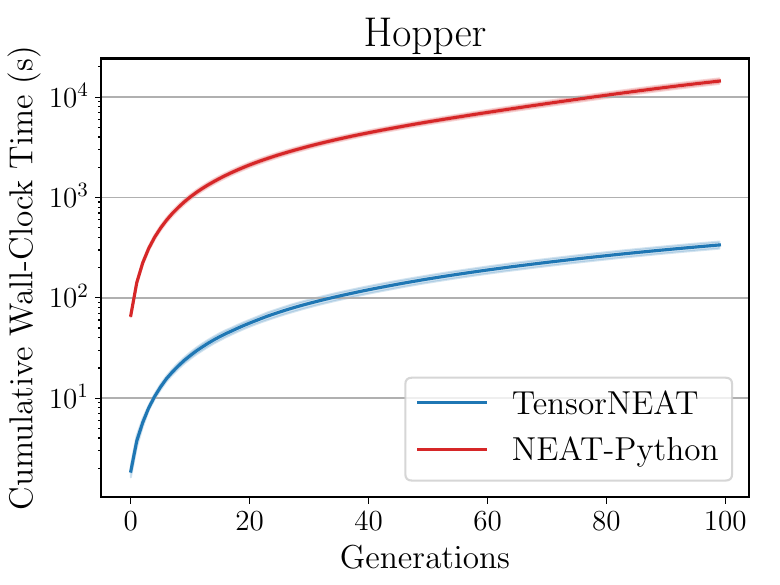}
    \end{subfigure}
    \hfill
    \begin{subfigure}{0.49\columnwidth}
        \centering
        \includegraphics[width=\columnwidth]{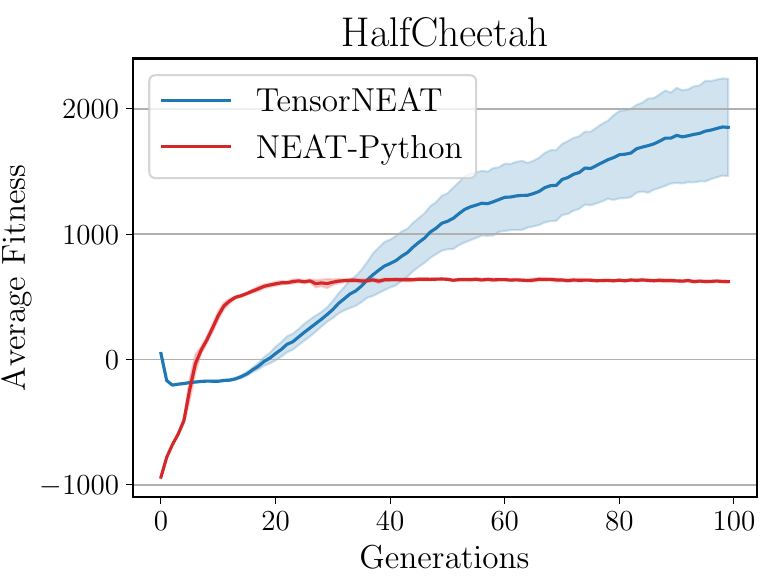}
    \end{subfigure}
    \hfill
    \begin{subfigure}{0.49\columnwidth}
        \centering
        \includegraphics[width=\columnwidth]{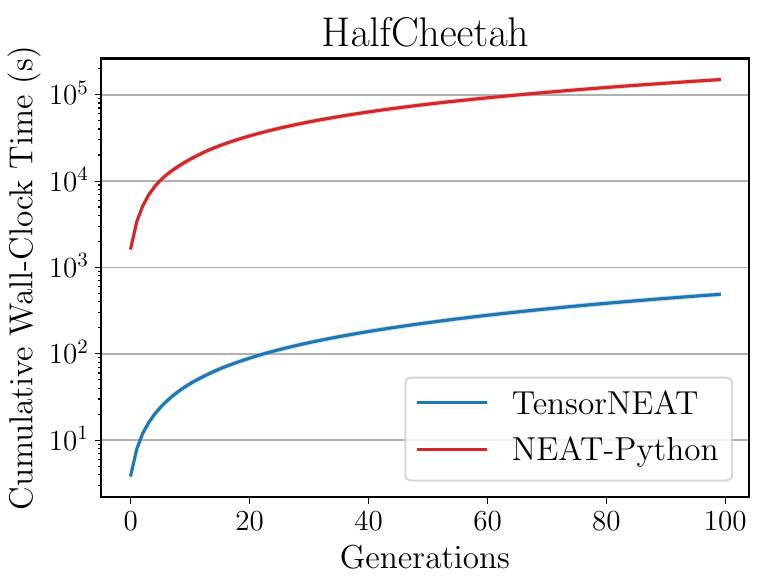}
    \end{subfigure}
    \caption{Average fitness and wall-clock time against generation.}
    \label{fig:fitness-generation}
\end{figure}

First, we examined the evolution of average population fitness and the cumulative runtime of the algorithms across generations, with a constant population size of 10,000. 
As depicted in Fig.~\ref{fig:fitness-generation}, TensorNEAT exhibits a more rapid improvement in population fitness over the course of the algorithm's execution. 
This performance disparity between the two frameworks stems from the modification in the NEAT algorithm after tensorization, including network encoding and the computation of distances between networks.
Furthermore, the analysis of execution times, illustrated in the final panel of Fig.~\ref{fig:fitness-generation}, reveals a significant decrease in runtime with TensorNEAT compared to NEAT-Python.

\begin{figure}[t]
    \centering
    \begin{subfigure}{0.49\columnwidth}
        \centering
        \includegraphics[width=\textwidth]{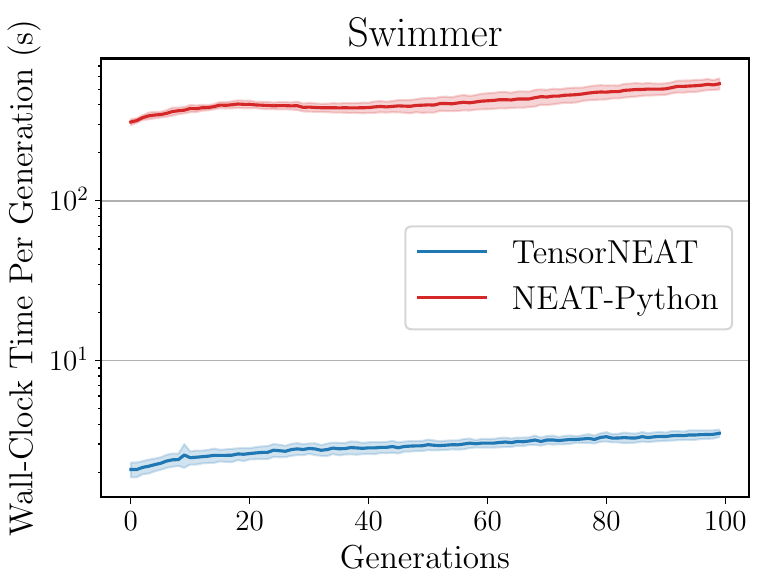}
    \end{subfigure}
    \hfill
    \begin{subfigure}{0.49\columnwidth}
        \centering
        \includegraphics[width=\textwidth]{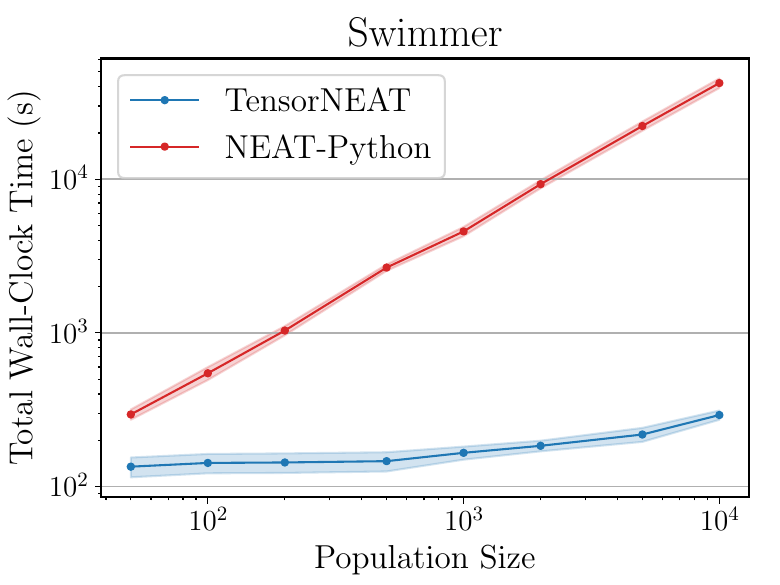}
    \end{subfigure}
    \hfill
    
    \begin{subfigure}{0.49\columnwidth}
        \centering
        \includegraphics[width=\textwidth]{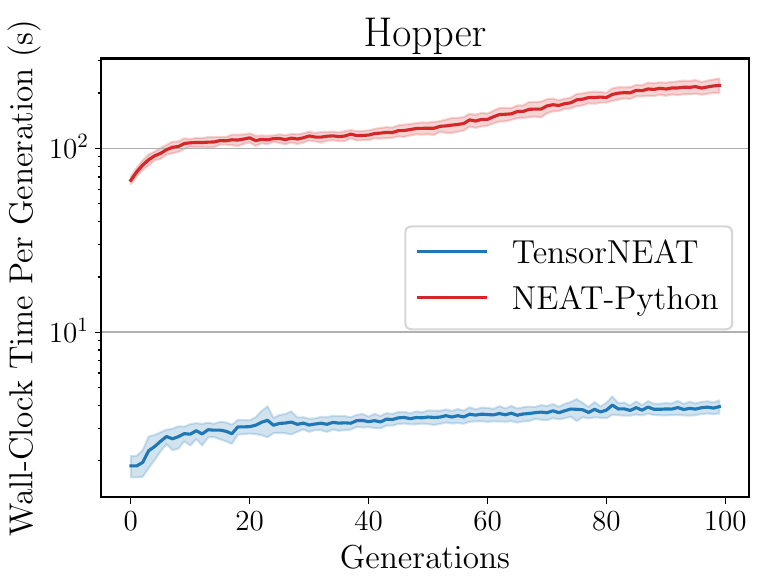}
    \end{subfigure}
    \hfill
        \begin{subfigure}{0.49\columnwidth}
        \centering
        \includegraphics[width=\textwidth]{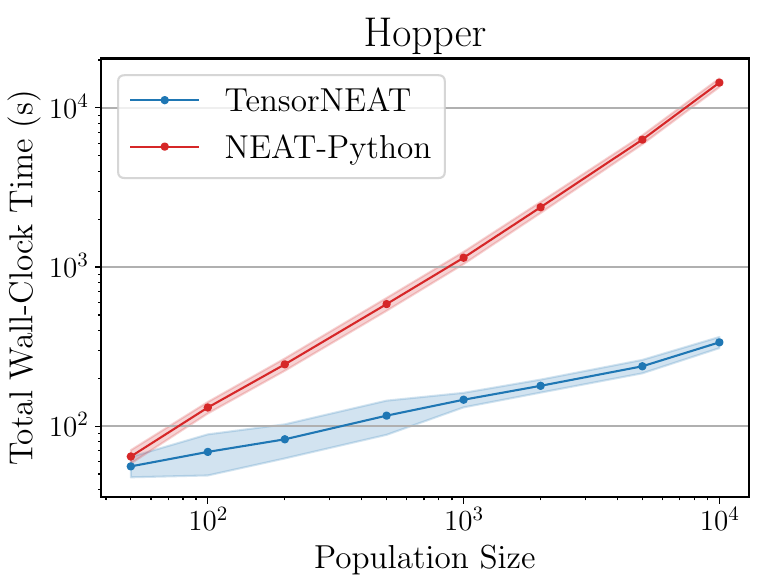}
    \end{subfigure}
    \hfill
    \begin{subfigure}{0.49\columnwidth}
        \centering
        \includegraphics[width=\textwidth]{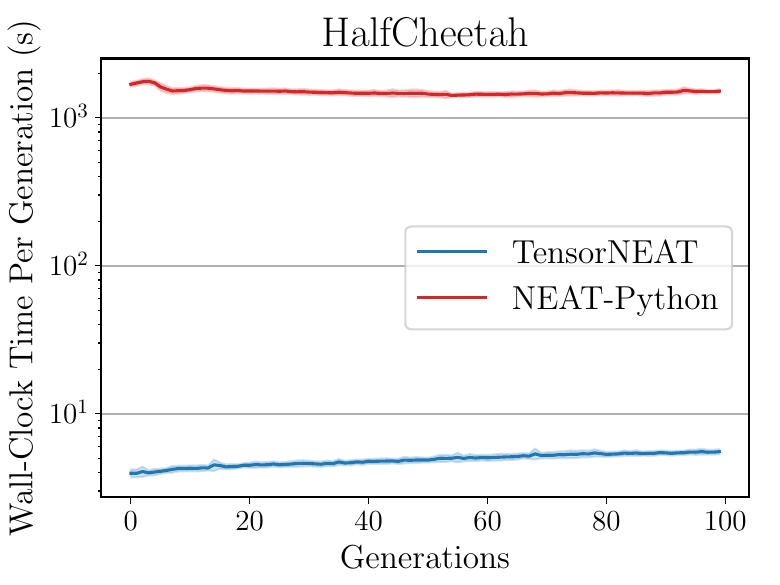}
    \end{subfigure}
    \hfill
     \begin{subfigure}{0.49\columnwidth}
        \centering
        \includegraphics[width=\textwidth]{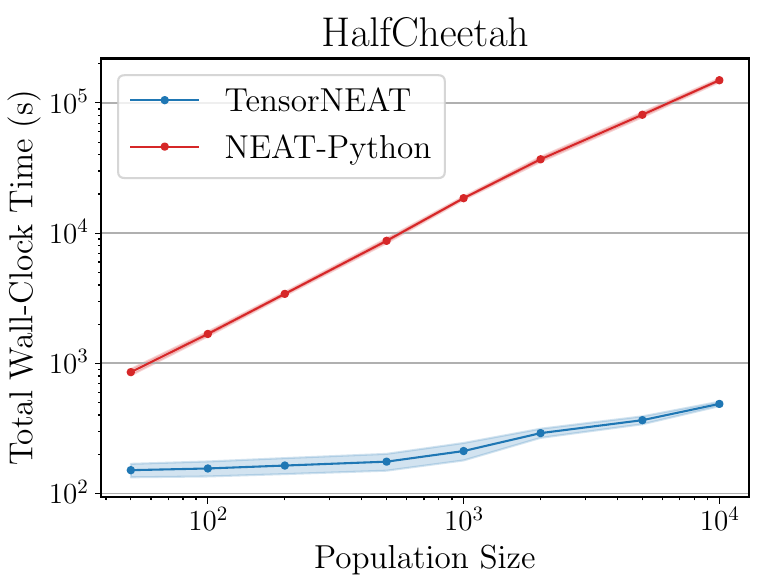}
    \end{subfigure}
    \caption{Wall-clock time against generation and population size. 
    }
    \label{fig:time-generation}
\end{figure}

Given the iterative nature of NEAT's process, there is an expected increase in per-iteration time as network structures become more complex. 
This phenomenon was investigated by comparing the per-generation runtimes of both algorithms. 
In the Swimmer and Hopper tasks, as shown in the left panels of Fig.~\ref{fig:time-generation}, NEAT-Python exhibits a more marked increase in runtime, likely due to its less efficient object encoding mechanism, which becomes increasingly cumbersome with rising network complexity. 
By contrast, TensorNEAT, employing a tensorized encoding approach constrained by the pre-set maximum values of $|N|_\text{max}$ and $|C|_\text{max}$, achieves a consistent network encoding size, resulting in more stable iteration times.

Additionally, we investigated how runtime varies with changes in population size, ranging from 50 to 10,000. 
As illustrated in the right panels of Fig.~\ref{fig:time-generation}, NEAT-Python's runtime significantly increases with larger populations, whereas TensorNEAT only experiences a marginal increase in runtime. 
TensorNEAT demonstrates advantages across various population sizes, with its benefits being more pronounced in larger populations.

The adaptability of TensorNEAT was further validated by evaluating its performance across various hardware configurations, including the AMD EPYC 7543 CPU and a selection of mainstream GPU models. 
These evaluations were conducted in the Cart Pole environment, with a consistent population size of $10,000$. 
The total wall-clock time was recorded over $100$ generations of the algorithm, and the results are summarized in Table~\ref{table:time_hardware}. 
In every GPU configuration, TensorNEAT demonstrated a significant performance advantage over NEAT-Python. Notably, with the RTX 4090, TensorNEAT achieved a speedup surpassing $500\times$ in the Halfcheetah environment.
It is also noteworthy that TensorNEAT realized a speedup on the same CPU device in comparison to NEAT-Python, especially in the more complex Halfcheetah environment.

Our experiments highlight TensorNEAT's considerable superiority to NEAT-Python in terms of algorithmic execution speed. 
Moreover, as the computational demands increase, whether due to larger network structures or higher population sizes, TensorNEAT's relative time consumption grows more slowly in contrast to NEAT-Python.
These findings emphasize the significant acceleration and performance efficiency of the TensorNEAT algorithm, especially with GPU accelerations.

\section{Conclusion}
In this work, we address the scalability challenge of NeuroEvolution of Augmenting Topologies (NEAT) by introducing a tensorization method, which transforms the network topologies into uniformly shaped tensors for efficient parallel processing.
Building upon the tensorization method, we develop TensorNEAT, which leverages JAX for automated function vectorization and hardware acceleration. 
Compatible with environments like Gym, Brax, and gymnax, TensorNEAT significantly outperforms traditional NEAT implementations, achieving over 500x speedups in various robotics control tasks. 

Looking ahead, our roadmap for TensorNEAT includes expanding its reach to distributed computing environments, transcending the limitations of single-machine setups. 
Furthermore, we plan to augment TensorNEAT's suite of functionalities by integrating advanced NEAT variants, such as DeepNEAT and CoDeepNEAT \cite{miikkulainen2019evolving}, to further enhance its potential in solving complex neuroevolution challenges.


\bibliographystyle{ACM-Reference-Format}
\bibliography{neatax-reference}

\appendix
\newpage
\onecolumn{}

\section{NeuroEvolution of Augmenting Topologies}\label{Appendix_neat}

\begin{algorithm}[h]
    \caption{Main Process of the NEAT algorithm}
    \label{alg:neat}
    \begin{algorithmic}
    \REQUIRE $P$ (population size), $I$ (number of input nodes), $O$ (number of output nodes), $f_{\text{target}}$ (target fitness value), $G$ (maximum number of generations) 
    \ENSURE $best$
    \STATE $pop \leftarrow$ initialize $P$ networks with $I$, $O$
    \FOR{$g = 1$ to $G$}
        \STATE $fit \leftarrow$ evaluate fitness values of $Pop$
        \IF{$\max(fit) \geq f_{\text{target}}$}
            \STATE \textbf{break}
        \ENDIF
        \STATE $species \leftarrow$ divide $pop$ by distances between networks
        
        \STATE $pop^* \leftarrow \{\}$ 
        \FOR{$s$ in $species$}
            \STATE $c \leftarrow$ determine the number of new individuals by $fit$
            \STATE $s \leftarrow$ generate $c$ networks using crossover and mutation
            \STATE $pop^* \leftarrow pop^* \cup s$
        \ENDFOR
        \STATE $pop \leftarrow pop^*$
    \ENDFOR
    \STATE \textbf{return} $pop[\arg\max(fit)]$
    \end{algorithmic}
\end{algorithm}

\newpage
\section{Hyperparameters}\label{Appendix_a}


The hyperparameters in TensorNEAT consists of those that influence the algorithm's dynamics and those who affect network behaviors:

\begin{itemize}
    \item Algorithmic Controls:
    \begin{itemize}
        \item \texttt{seed}: Random seed (integer).
        \item \texttt{fitness\_target}: Target fitness value for termintaion (float).
        \item \texttt{generation\_limit}: Maximum number of generations for termintaion (float).
        \item \texttt{pop\_size}: Population size (integer).
        \item \texttt{network\_type}: Network type, either \texttt{feedforward} (no cycles) or \texttt{recurrent} (with cycles).
        \item \texttt{inputs}: Number of network inputs (integer).
        \item \texttt{outputs}: Number of network outputs (integer).
        \item \texttt{max\_nodes}: Max nodes allowed in a network (integer).
        \item \texttt{max\_conns}: Max connections allowed in a network (integer).
        \item \texttt{max\_species}: Max species in the population (integer).
        \item \texttt{compatibility\_disjoint}: Weight for disjoint genes in distance calculation between genomes (float).
        \item \texttt{compatibility\_homologous}: Weight for homologous genes in distance calculation between genomes (float).
        \item \texttt{node\_add}: Probability of a node addition during mutation (float).
        \item \texttt{node\_delete}: Probability of a node deletion during mutation (float).
        \item \texttt{conn\_add}: Probability of a connection addition during mutation (float).
        \item \texttt{conn\_delete}: Probability of a connection deletion during mutation (float).
        \item \texttt{compatibility\_threshold}: Distance threshold for genome speciation (float). 
        \item \texttt{species\_elitism}: Minimum species count to prevent all species are stagnated (integer).
        \item \texttt{max\_stagnation}: Stagnation threshold for species (integer). If a species does not show improvement for \texttt{max\_stagnation} consecutive generations, then this species will be stagnated.
        \item \texttt{genome\_elitism}: Number of elite genomes preserved for next generation (integer).
        \item \texttt{survival\_threshold}: Percentage of species survival for crossover (float).
        \item \texttt{spawn\_number\_change\_rate}: Rate of change in species size over two consecutive generations (float).
    \end{itemize}

    \item Network Behavior Controls:
    \begin{itemize}
        \item \texttt{bias\_init\_mean}: Mean value for bias initialization (float).
        \item \texttt{bias\_init\_std}: Standard deviation for bias initialization (float).
        \item \texttt{bias\_mutate\_power}: Mutation strength for bias values (float).
        \item \texttt{bias\_mutate\_rate}: Probability of bias value mutation (float).
        \item \texttt{bias\_replace\_rate}: Probability to replace existing bias with a new value during mutation (float).

        \item \texttt{response\_init\_mean}: Mean value for response initialization (float).
        \item \texttt{response\_init\_std}: Standard deviation for response initialization (float).
        \item \texttt{response\_mutate\_power}: Mutation strength for response values (float).
        \item \texttt{response\_mutate\_rate}: Probability of response value mutation (float).
        \item \texttt{response\_replace\_rate}: Probability to replace existing response with a new value during mutation (float).

        \item \texttt{weight\_init\_mean}: Mean value for weight initialization (float).
        \item \texttt{weight\_init\_std}: Standard deviation for weight initialization (float).
        \item \texttt{weight\_mutate\_power}: Mutation strength for weight values (float).
        \item \texttt{weight\_mutate\_rate}: Probability of weight value mutation (float).
        \item \texttt{weight\_replace\_rate}: Probability to replace existing weight with a new value during mutation (float).

        \item \texttt{activation\_default}: Default value for activation function.
        \item \texttt{activation\_options}: Available activation functions.
        \item \texttt{activation\_replace\_rate}: Probability to change the activation function during mutation.

        \item \texttt{aggregation\_default}: Default value for aggregation function.
        \item \texttt{aggregation\_options}: Available aggregation functions.
        \item \texttt{aggregation\_replace\_rate}: Probability to change the aggregation function during mutation.
    \end{itemize}
\end{itemize}

\newpage

\section{Interfaces}\label{Appendix_b}

\subsection{Network Interface}

TensorNEAT offers users the flexibility to define a custom network that will be optimized by the NEAT algorithms by providing a network interface (as presented in Listing~\ref{lst:gene_interface} and Listing~\ref{lst:genome_interface}). Users are required to define the specific behaviors associated with their network, including initialization, mutation, distance computation, and inference.

By implementing these interfaces, users can fit their specific requirements and leverage TensorNEAT's power for a wide range of neural network architectures.

\begin{figure}[H]
\begin{lstlisting}[
language=Python,
caption=Gene interface. ,
label=lst:gene_interface
]
class BaseGene:
    fixed_attrs = []
    custom_attrs = []

    def __init__(self):
        pass

    def new_custom_attrs(self):
        raise NotImplementedError

    def mutate(self, randkey, gene):
        raise NotImplementedError

    def distance(self, gene1, gene2):
        raise NotImplementedError

    def forward(self, attrs, inputs):
        raise NotImplementedError

    @property
    def length(self):
        return len(self.fixed_attrs) + len(self.custom_attrs)
\end{lstlisting}
\end{figure}

\begin{figure}[H]
\begin{lstlisting}[
language=Python,
caption=Genome interface. ,
label=lst:genome_interface
]
class BaseGenome:
    network_type = None

    def __init__(
            self,
            num_inputs: int,
            num_outputs: int,
            max_nodes: int,
            max_conns: int,
            node_gene: BaseNodeGene = DefaultNodeGene(),
            conn_gene: BaseConnGene = DefaultConnGene(),
    ):
        self.num_inputs = num_inputs
        self.num_outputs = num_outputs
        self.input_idx = jnp.arange(num_inputs)
        self.output_idx = jnp.arange(num_inputs, num_inputs + num_outputs)
        self.max_nodes = max_nodes
        self.max_conns = max_conns
        self.node_gene = node_gene
        self.conn_gene = conn_gene

    def transform(self, nodes, conns):
        raise NotImplementedError

    def forward(self, inputs, transformed):
        raise NotImplementedError
\end{lstlisting}
\end{figure}

\newpage

\subsection{Problem Template}

In TensorNEAT, users also have the flexibility to define custom problems they wish to optimize using the NEAT algorithms. This can be done by implementing the \texttt{Problem} interface as illustrated in Alg.~\ref{lst:problem_interface}. Within this interface, users are required to provide the evaluation process for the problem, specify the input and output dimensions, and optionally implement a function to visualize the solution.

\begin{figure}[H]
\begin{lstlisting}[
language=Python,
caption=Problem interface in TensorNEAT. ,
label=lst:problem_interface
]
from typing import Callable

from utils import State


class BaseProblem:
    jitable = None

    def setup(self, randkey, state: State = State()):
        """initialize the state of the problem"""
        raise NotImplementedError

    def evaluate(self, randkey, state: State, act_func: Callable, params):
        """evaluate one individual"""
        raise NotImplementedError

    @property
    def input_shape(self):
        raise NotImplementedError

    @property
    def output_shape(self):
        raise NotImplementedError

    def show(self, randkey, state: State, act_func: Callable, params, *args, **kwargs):
        raise NotImplementedError

\end{lstlisting}
\end{figure}

\newpage
\section{Experiment Detail}\label{Appendix_c}
In this section, we detail the environments and hyperparameters in experiments.
\subsection{Environments}
\begin{figure}[htbp]
    \begin{subfigure}{0.20\columnwidth}
        \centering
        \includegraphics[width=\columnwidth]{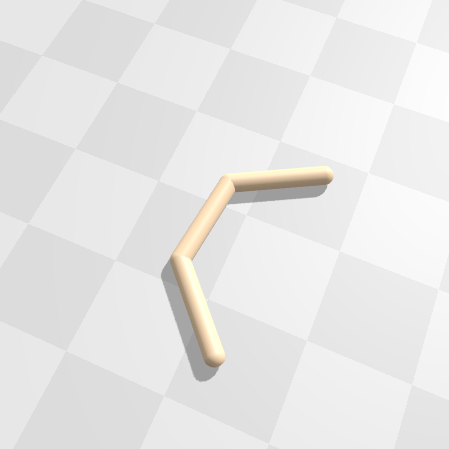}
        \caption{Swimmer}
        \label{fig:swimmer}
    \end{subfigure}
    \begin{subfigure}{0.20\columnwidth}
        \centering
        \includegraphics[width=\columnwidth]{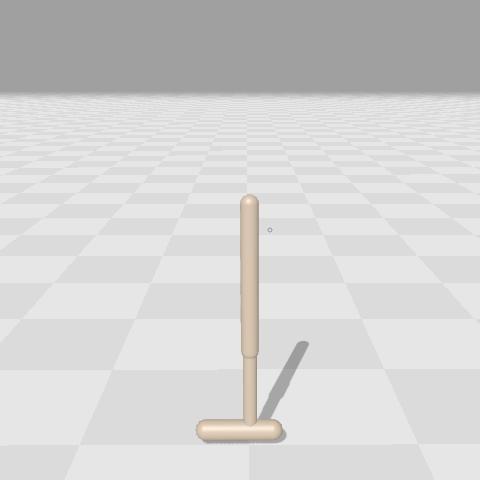}
        \caption{Hopper}
        \label{fig:hopper}
    \end{subfigure}
    \begin{subfigure}{0.20\columnwidth}
        \centering
        \includegraphics[width=\columnwidth]{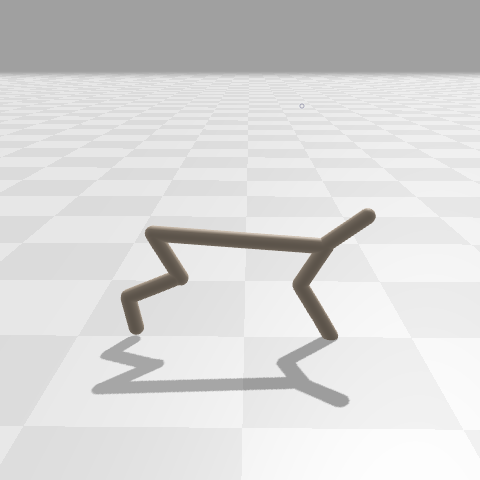}
        \caption{Halfcheetah}
        \label{fig:halfcheetah}
    \end{subfigure}
    \caption{Robotics Control tasks in Brax.}
    \label{fig:tasks}
\end{figure}

We compare TensorNEAT to NEAT-Python in three reinforcement learning environments in Brax: Swimmer, Hopper and Halfcheetah. They are robotic control tasks and fig~\ref{fig:tasks} are screenshots of them. The primary objective of the robots is to advance as far as possible in a forward direction.
These three environments have different action dimension and observation dimension, which are shown in table~\ref{table:experiment settings}.

\begin{table}[htbp]
    \centering
    \begin{tabular}{|c|c|c|}
            
            \hline
            Environments & Action Dimension & Observertion Dimension\\
            \hline
            Swimmer & 2 & 8 \\
            \hline
            Hopper & 3 & 11 \\
            \hline
            Halfcheetah & 6 & 18 \\
       
            \hline
    \end{tabular}
    \caption{Action dimension and observation dimension of environments.}
    \label{table:experiment settings}
\end{table}

\subsection{Hyperparameters}

Hyperparameters in TensorNEAT:
\begin{itemize}
    \item Algorithmic Controls:
    \begin{itemize}
        \item \texttt{seed}: \texttt{[0, 1, 2, 3, 4, 5, 6, 7, 8, 9]};
        \item \texttt{fitness\_target}: \texttt{Inf (to terminate at the fixed generation)};
        \item \texttt{generation\_limit}: \texttt{100};
        \item \texttt{pop\_size}: \texttt{[50, 100, 200, 500, 1000, 2000, 5000, 10000]};
        \item \texttt{network\_type}: \texttt{feedforward};
        \item \texttt{inputs}: the same as the observation dimension of the environment.
        \item \texttt{outputs}: the same as the action dimension of the environment.
        \item \texttt{max\_nodes}: \texttt{50};
        \item \texttt{max\_conns}: \texttt{100};
        \item \texttt{max\_species}: \texttt{10};
        \item \texttt{compatibility\_disjoint}: \texttt{1.0};
        \item \texttt{compatibility\_homologous}: \texttt{0.5};
        \item \texttt{node\_add}: \texttt{0.2};
        \item \texttt{node\_delete}: \texttt{0};
        \item \texttt{conn\_add}: \texttt{0.4};
        \item \texttt{conn\_delete}: \texttt{0};
        \item \texttt{compatibility\_threshold}: \texttt{3.5};
        \item \texttt{species\_elitism}: \texttt{2};
        \item \texttt{max\_stagnation}: \texttt{15};
        \item \texttt{genome\_elitism}: \texttt{2};
        \item \texttt{survival\_threshold}: \texttt{0.2};
        \item \texttt{spawn\_number\_change\_rate}: \texttt{0.5};
    \end{itemize}

    \item Network Behavior Controls:
    \begin{itemize}
        \item \texttt{bias\_init\_mean}: \texttt{0};
        \item \texttt{bias\_init\_std}: \texttt{1.0};
        \item \texttt{bias\_mutate\_power}: \texttt{0.5};
        \item \texttt{bias\_mutate\_rate}: \texttt{0.7};
        \item \texttt{bias\_replace\_rate}: \texttt{0.1};

        \item \texttt{response\_init\_mean}: \texttt{1.0};
        \item \texttt{response\_init\_std}: \texttt{0};
        \item \texttt{response\_mutate\_power}: \texttt{0};
        \item \texttt{response\_mutate\_rate}: \texttt{0};
        \item \texttt{response\_replace\_rate}: \texttt{0};

        \item \texttt{weight\_init\_mean}: \texttt{0};
        \item \texttt{weight\_init\_std}: \texttt{1};
        \item \texttt{weight\_mutate\_power}: \texttt{0.5};
        \item \texttt{weight\_mutate\_rate}: \texttt{0.8};
        \item \texttt{weight\_replace\_rate}: \texttt{0.1};

        \item \texttt{activation\_default}: \texttt{tanh};
        \item \texttt{activation\_options}: \texttt{[tanh]};
        \item \texttt{activation\_replace\_rate}: \texttt{0};

        \item \texttt{aggregation\_default}: \texttt{sum};
        \item \texttt{aggregation\_options}: \texttt{[sum]};
        \item \texttt{aggregation\_replace\_rate}: \texttt{0};
    \end{itemize}
\end{itemize}

Hyperparameters in NEAT-Python:
\begin{itemize}
    \item \texttt{fitness\_criterion}: \texttt{max}
    \item \texttt{fitness\_threshold}: \texttt{999}
    \item \texttt{pop\_size}: \texttt{10000}
    \item \texttt{reset\_on\_extinction}: \texttt{False}

    \item \textbf{[DefaultGenome]}
    \begin{itemize}
        \item \texttt{activation\_default}: \texttt{tanh}
        \item \texttt{activation\_mutate\_rate}: \texttt{0}
        \item \texttt{activation\_options}: \texttt{tanh}
        \item \texttt{aggregation\_default}: \texttt{sum}
        \item \texttt{aggregation\_mutate\_rate}: \texttt{0.0}
        \item \texttt{aggregation\_options}: \texttt{sum}
        \item \texttt{bias\_init\_mean}: \texttt{0.0}
        \item \texttt{bias\_init\_stdev}: \texttt{1.0}
        \item \texttt{bias\_max\_value}: \texttt{30.0}
        \item \texttt{bias\_min\_value}: \texttt{-30.0}
        \item \texttt{bias\_mutate\_power}: \texttt{0.5}
        \item \texttt{bias\_mutate\_rate}: \texttt{0.7}
        \item \texttt{bias\_replace\_rate}: \texttt{0.1}
        \item \texttt{compatibility\_disjoint\_coefficient}: \texttt{1.0}
        \item \texttt{compatibility\_weight\_coefficient}: \texttt{0.5}
        \item \texttt{conn\_add\_prob}: \texttt{0.5}
        \item \texttt{conn\_delete\_prob}: \texttt{0}
        \item \texttt{enabled\_default}: \texttt{True}
        \item \texttt{enabled\_mutate\_rate}: \texttt{0.01}
        \item \texttt{feed\_forward}: \texttt{True}
        \item \texttt{initial\_connection}: \texttt{full}
        \item \texttt{node\_add\_prob}: \texttt{0.2}
        \item \texttt{node\_delete\_prob}: \texttt{0}
        \item \texttt{num\_hidden}: \texttt{0}
        \item \texttt{num\_inputs}: the same as the observation dimension of the environment.
        \item \texttt{num\_outputs}: the same as the action dimension of the environment.
        \item \texttt{response\_init\_mean}: \texttt{1.0}
        \item \texttt{response\_init\_stdev}: \texttt{0.0}
        \item \texttt{response\_max\_value}: \texttt{30.0}
        \item \texttt{response\_min\_value}: \texttt{-30.0}
        \item \texttt{response\_mutate\_power}: \texttt{0.0}
        \item \texttt{response\_mutate\_rate}: \texttt{0.0}
        \item \texttt{response\_replace\_rate}: \texttt{0.0}
        \item \texttt{weight\_init\_mean}: \texttt{0.0}
        \item \texttt{weight\_init\_stdev}: \texttt{1.0}
        \item \texttt{weight\_max\_value}: \texttt{30}
        \item \texttt{weight\_min\_value}: \texttt{-30}
        \item \texttt{weight\_mutate\_power}: \texttt{0.5}
        \item \texttt{weight\_mutate\_rate}: \texttt{0.8}
        \item \texttt{weight\_replace\_rate}: \texttt{0.1}
    \end{itemize}

    \item \textbf{[DefaultSpeciesSet]}
    \begin{itemize}
        \item \texttt{compatibility\_threshold}: \texttt{3.0}
    \end{itemize}

    \item \textbf{[DefaultStagnation]}
    \begin{itemize}
        \item \texttt{species\_fitness\_func}: \texttt{max}
        \item \texttt{max\_stagnation}: \texttt{20}
        \item \texttt{species\_elitism}: \texttt{2}
    \end{itemize}

    \item \textbf{[DefaultReproduction]}
    \begin{itemize}
        \item \texttt{elitism}: \texttt{-9999}
        \item \texttt{survival\_threshold}: \texttt{0.2}
    \end{itemize}
\end{itemize}

\end{document}